\documentclass[runningheads]{llncs}
\usepackage{graphicx}
\usepackage{comment}
\usepackage{amsmath,amssymb,bbm} % define this before the line numbering.
\usepackage{caption}
\usepackage{subcaption}
\usepackage{color}
\usepackage{wrapfig}

\begin{document}
% \linenumbers
\pagestyle{headings}
\mainmatter
\def\ECCVSubNumber{16}  % Insert your submission number here

\title{Learning Portrait Style Representations} % Replace with your title

\begin{comment}
\titlerunning{ECCV-20 submission ID \ECCVSubNumber} 
\authorrunning{ECCV-20 submission ID \ECCVSubNumber} 
\author{Anonymous ECCV submission}
\institute{Paper ID \ECCVSubNumber}
\end{comment}
%******************

%\begin{comment}
\titlerunning{Learning Portrait Style Representations}
\author{Sadat Shaik\thanks{Equal contribution.}\inst{1} \and Bernadette Bucher$^{\star}$\inst{1}
Nephele Agrafiotis\inst{1}\and Stephen Phillips\inst{1}
Kostas Daniilidis\inst{1} \and William Schmenner\inst{1}}
\authorrunning{S. Shaik et al.}
\institute{University of Pennsylvania, Philadelphia PA \\
\email{\{sshaik, bucherb\}@seas.upenn.edu}}
%\end{comment}
%******************
\maketitle
% 150 words max
\begin{abstract}
Style analysis of artwork in computer vision predominantly focuses on achieving results in target image generation through optimizing understanding of low level style characteristics such as brush strokes. However, fundamentally different techniques are required to computationally understand and control qualities of art which incorporate higher level style characteristics. We study style representations learned by neural network architectures incorporating these higher level characteristics. We find variation in learned style features from incorporating triplets annotated by art historians as supervision for style similarity. Networks leveraging statistical priors or pretrained on photo collections such as ImageNet can also derive useful visual representations of artwork. We align the impact of these expert human knowledge, statistical, and photo realism priors on style representations with art historical research and use these representations to perform zero-shot classification of artists. To facilitate this work, we also present the first large-scale dataset of portraits prepared for computational analysis.

\keywords{style analysis, representation learning, triplet loss, deep metric learning, explainable AI, generative models}
\end{abstract}

%%%%%%%%% BODY TEXT
\section{Introduction}%TODO citations and figures

% Style transfer in computer vision isolates components of image style from the content of an image so that style information can then be used to reconstruct images with different content but the same style. Current methods for style transfer disentangle content of the scenes from style. Style is sometimes also further disentangled into components such as stroke width and color.

Interdisciplinary applications across computer science and the visual arts motivate research on style in computer vision. However, most of the existing literature in computer vision has not been pursued as an interdisciplinary effort with experts across these fields. As a result, the definitions of style exploited by existing methods frequently do not directly align with definitions readily accepted by art historians. Furthermore, expanding the scope of style research to perform more fine-grained and technically precise analysis of artistic properties is challenging due to the the lack of appropriate datasets \cite{Jing2017NeuralST}.

In most computer vision research, style is defined separately from image content in a manner which closely relates style to local features like image texture. Typically, image content is defined by a photograph which is considered to be the base representation of content and treated as being without style. Then, a texture synthesis procedure is used on a single painting, and the texture from that painting is applied to a target photograph in a manner which preserves content such that it is still clearly human recognizable \cite{firstOne}. Furthermore, this texture synthesis of paintings relies on the existence of easily recognizable brushstrokes which is prevalent primarily in 19th to early 20th century artwork biasing computer vision research toward computational analyses of style tailored to specific subsets of artistic style and art history. The few readily available datasets supporting computational style research add to these biases \cite{data,Shen_2019_CVPR}.

More recently, style representations have been proposed by the vision community in which a space of styles is derived from multiple images and synthesized in a set which can either be interpolated across or arbitrarily combined \cite{interpretableStyle,gatedGAN,learnedStyleRepresentation,sanakoyeu2018styleaware}. One recent approach even incorporates scene content as a part of style \cite{kotovenko2019cvpr}. Thus, the predominant research trends in the study of style in computer vision is providing computational analysis and definitions of style which align more closely with art historical analysis.

Art historians both derive understanding of style from collections of artwork and consider content as a part of style. Style is considered historically bounded, for example, by the materials available for creating art at any given time, by the networks and infrastructures available to educate and employ artists, and by the objects in the world. Artists painting in the 19th and early 20th century, the dominant type of artist considered by most style transfer studies, could and did chose what they painted. Paul Cezanne choosing to paint Mount Sainte Victoire or the card players, says something as significant about his style as how he paints. The Impressionists decision to paint \textit{en plein air} affected both how they painted, more quickly, and what they painted, the places of leisure of a modernizing Paris, for instance. Style is more than the quality, including texture, of individual brush strokes, and this is especially evident in early oil painting, when many artists, such as Jan van Eyck, aspired to hide their brush strokes entirely. For these and other reasons, style cannot apply only to easily spotted brush strokes from certain artists working in the 19th to early 20th century.

%Some existing computer vision research has addressed art and computer vision questions jointly \cite{Shen_2019_CVPR} \cite{bourached2019raiders} 

We consider these facets of style in order to take a step toward enabling modern computer vision techniques to develop a computational understanding of artist style which aligns with professional human interpretation and analysis. To support this goal, we provide the following contributions in this work.

\begin{enumerate}
    \item \textbf{Professionally curated dataset of portraits.} We present the first large-scale dataset of portrait paintings prepared for computational analysis. We include time period and artist metadata as well as professionally annotated style similarity triplet labels.
    \item \textbf{Style representation analysis.} We investigate whether established computer vision models are adequate representations for art style. We propose a generative model that utilizes style annotations from art historians during training. When this model is applied to pre-learned computer vision representations, it yields embeddings whose relations align most closely with that of a human expert. Through this learned style we also show that this approach is able to perform zero-shot artist classification.
    % In this analysis, we demonstrate that the use of priors provided by human labels and photo collections in additional to model architecture decisions create variations in learned features which correspond to different aspects of art historical analysis of style.
\end{enumerate}

\section{Related Work}

% TODO - should we make a section on datasets?
%\textbf{Datasets.}\cite{data} \cite{Shen_2019_CVPR}

\textbf{Single Image Style Analysis and Transfer.} A number of methods exist to disentangle a representation of style from art for use in transferring this style to photographic images. \cite{firstOne} uses losses in intermediate feature representations to partition texture from content in images, and image texture is equated with style. The extracted texture from a single piece of art is overlaid on a photograph to complete the process of style transfer. Several extensions to this work improve results in various ways using a similar base architecture. \cite{depthPreservingStyleTransfer} incorporates depth maps preservation from content images. \cite{arbitraryStyleTransfer} aligns means and variances of texture and content information in the transfer process. \cite{Yao_2019_CVPR} uses attention mechanisms for emphasizing multi-scale texture synthesis. \cite{Johnson2016Perceptual} defines network perceptual losses to improve transfer resolution and speed. \cite{strokeControllableStyle} provides a method for transferring brush stroke texture at varying scales to a content image. \cite{universalStyle} incorporates whitening and coloring transforms in the image reconstruction network to preserve a direct matching of feature covariance between the style and content images.

Some style transfer literature has deviated more strongly from the initial approach in \cite{firstOne}. \cite{selim16} defined style by local color distributions which showed to be successful in transferring style information to head portrait photographs where detailed geometric information needed to be retained in order to recognize the subjects in the photographs. \cite{Azadi2017MulticontentGF} uses a GAN architecture to define content by more than a single photographic image. \cite{kotovenko2019iccv} presents a new approach for disentangling content from style while also introducing a triplet loss approach to understanding shape variation as a component of style. In contrast to these methods, our analysis focuses on learning representations of style decomposed from a collection of images.

\textbf{Multi-Image Style Analysis and Transfer.} Recent work in style transfer has developed style representations in which styles from multiple images are mapped into a continuous embedding on which interpolation can be performed. \cite{interpretableStyle} decomposes styles onto a basis for a continuous vector space so that arbitrary combinations of styles can be constructed from this basis and overlaid onto arbitrary content images. \cite{learnedStyleRepresentation} also learns a lower dimensional embedding of style across multiple images which can be used to overlay content images given by photographs. Adversarial gated networks in \cite{gatedGAN} enable different style components to be synthesized and combined in style transfer. \cite{sanakoyeu2018styleaware} introduces a style-aware content loss to the single image style transfer framework so that style can be learned across multiple images and transferred to images with varying content.

The approach in \cite{kotovenko2019cvpr} learns style representations most similar to those we analyze here due to their treatment of content as a part of style with a comparable goal of remaining consistent with art historical analysis. However, content is partitioned as a distinct component of style and explicitly transformed during style transfer experiments which still incorporate the base content from a target scene for use in stylization. In contrast, we do not partition content from the style representations we analyze.

\textbf{Style-Conditioned Generative Models.} \cite{artGAN,artgan2018} demonstrate the ability to generate novel artwork in the style of a predetermined class label such as a specific genre and artist from the set of labels specified in training. \cite{styleGAN} proposes an approach to control scale-specific style in generated images of photographs of human faces. In \cite{Gilbert_2018_CVPR}, partial image generation is accomplished via style-aware image completion for which a comparable approach to style transfer methods is used to disentangle structure from style where style is treated as analogous to image texture. These models do not learn distinct representations of existing artwork. However, further analyzing the feature responses given by generated artwork from these models is an interesting direction for future research.

\textbf{Deep Metric Learning.} We incorporate expert human knowledge as a model prior in our analysis through the adaption existing deep metric learning techniques. Given the extensive literature on deep metric learning, we only highlight a selection of papers in the field here. Metric learning approaches optimize learning a model parameterization by enforcing sample similarity through loss functions. Deep metric learning refers to metric learning optimization procedures for neural networks. Deep metric learning approaches are primarily achieved with either triplet losses \cite{tripletLoss} or Siamese networks \cite{bromley1994signature}. These approaches to learning sample similarity are effectively used to accomplish many computer tasks including image clustering \cite{sohn2016improved,wang2017deep,oh2016deep,oh2017deep,ge2018deep},
person reidentification \cite{ustinova2016learning,chen2018person,yang2018person,yao2019deep}, 3D shape retrival \cite{dai2017deep,dai2018deep,he2018triplet}, and face verification \cite{hu2014discriminative,lu2017discriminative,hu2017sharable,Schroff2015}. Deep metric learning has also shown usefulness in natural language processing tasks including identifying semantic textual similarity \cite{mueller2016siamese,zhu2018dependency,dorlearning} and speaker verification \cite{wang2019centroid,chen2019distance,hoffer2015deep}. In this work, we incorporate a triplet loss in a VAE to enforce style similarity defined by expert provided labels.

\textbf{Representation Learning.} Representation learning is widely studied across many computational fields including computer vision. We only provide a review of some of the key representation learning papers in computer vision here due to the expansiveness of this topic. Learning partitioned feature representations in the latent space of networks has been previously used for tasks requiring geometric understanding. \cite{disentangle} extracts 3D pose from multi-view images using a network which disentangles latent 3D pose and geometry, appearance, and background information. In \cite{esser2019unsupervised}, disentangled latent representations of object pose and appearance are learned without supervision by controlling for appearance and varying pose in the image pairs used for training. \cite{scooped,BucherCVPR2019} both learn a latent representation of geometry separate from appearance for 2D images for the purpose of depth recovery and novel view synthesis. A general formulation of this latent class-based partitioning for generative models was proposed in \cite{ilse2019diva}. In this work, the connections we make between computational to art historical style analysis means to drive research in learning useful representations for style which can tie more closely to meaningful style characteristics studied for centuries by art historians.

\textbf{Our Contributions in Context} We demonstrate trade-offs in the information incorporated in different representations of style for analysis of art data. The closest works to our own are \cite{elgammal2018shape} and \cite{kotovenko2019cvpr}. \cite{elgammal2018shape} uses a careful application of Wölfflin’s concepts to also analyze representations of style. In contrast, we choose a continuous style representation where style interpretation from art historians is included through supervision from triplet labels. Similarly to \cite{elgammal2018shape}, we start with pre-learned representation, but we use a generative model with a triplet loss for style similarity instead of a style classification. While \cite{kotovenko2019cvpr} also design a generative model which takes into account art historical definitions of style, they do not explicitly analyze how network priors and architectures align with art historical concepts of style in different ways. We are the first work to perform an interdisciplinary analysis of style representations in this manner.

The style representations we consider use neural network architectures which learn from data of many images to construct the representations we study. We also learn style holistically from images without disentangling content information in alignment with art historical analysis techniques. The style representations we study are derived from digital scans of human-made portrait paintings for which we collected and annotated a dataset. In addition to varying network architectures, we also demonstrate representation changes when different priors are used, particularly from photographs and expert human knowledge. To incorporate expert human knowledge as a network prior for our representation analysis, we use a triplet loss to enforce style similarity based on labels provided by a professional art historian. The results of our analysis can be used to incorporate a more expansive set of knowledge into future models manipulating and decoding artistic style as well as provide additional insight to art historians both about human-made artwork and the new trend of machine-generated artwork using neural networks.

\section{The Portrait Dataset}

To build our dataset of portraits, we scraped 6886 images from Artstor \cite{artstor}, an open source digital database of artwork. The data provided via this website and similar digital art collections is not initially well-suited to computational analysis. Challenges for using this data for computational analysis include labels which are not unique or do not have a standard format such as artist name labels and date formatting. These metadata fields needed to be cleaned for easy computational processing.

%We chose portraits restricted to European and North American origins as our target content for several reasons. First, we wanted to restrict content to a broad category due to the extremely wide variation in style including subjects across artwork. To perform a first detailed analysis of style representations, it was necessary to scope our analyzed content in order to be able to make meaningful comparisons and conclusions from our variation in computational approaches. In addition, the accessibility of digital artwork from Europe and North America was much greater than that from other geographic locations. To include content from different geographical regions would include very distinct style examples. Since they would be underrepresented styles in our data collection, we anticipated that incorporating this style diversity in an unbalanced way would confound our results.

In addition to prepared metadata for artists and time period, we also provide triplet labels for all the images. Each image in the dataset is an anchor for one triplet. The other two images in the triplet are randomly assigned to this grouping. A professional art historian marked for each of these triplets which image was closer in style to the anchor image. This more stylistically similar image is the "positive" example and the other is the "negative" example. We discuss the use of these labels further in our explanation of our methods which use this data. We built a publicly available website\footnote{https://bucherb.github.io/triplet-labels/} for creating these triplet labels which can be used as a tool to build triplet labels for any dataset.

Prior to our collection, the only publicly available datasets of digital artwork curated for computational analysis of which we are aware are \cite{Shen_2019_CVPR} and \cite{kaggle}. To perform a first detailed analysis of style representations, it was necessary to scope our analyzed content in order to be able to make meaningful comparisons and conclusions from our computational approaches. The diverse content in both of these datasets yet more limited variations in time period and artist which contributes more strongly to style differences led us to believe that they would not suite the purpose of our analysis as well. Given the varied requirements of computational analysis of artwork, more digital collections with parsable metadata and varied types of human annotations are required to continue expanding this area of research.

Our dataset is available at \url{https://nephagra.github.io/portrait-dataset-submission/}.

\section{Model Priors}\label{sec:priors}

% What are the key results we have? VGG-16 + triplet loss (human and photo prior), VGG-16 (photo prior only), VAE + triplet loss (human + statistical priors), VAE (statistical prior only)

We incorporate three different priors in our networks to study their impact on building meaningful style representations for our portrait dataset.

\begin{enumerate}
    \item \textbf{An expert human knowledge prior} is enforced through a style similarity triplet loss for our portrait data given by triplet annotations from a professional art historian.
    \item \textbf{A statistical prior} is enforced through the use of a variational autoencoder (VAE) \cite{Kingma2014} trained on our portrait dataset only.
    \item \textbf{A photo realism prior} is enforced through the use of a VGG-16 network trained on ImageNet \cite{deng2009imagenet}.
\end{enumerate}

We hypothesize that these priors will capture different aspects of style, some of which will result in more meaningful visual representations than others. In particular, we expect that the photo realism prior will enable feature extraction of low and intermediate level style characteristics. We further hypothesize that the statistical prior imposed by our VAE will allow a reasonable embedding of style from our portrait dataset by enforcing distributional properties when the space we are trying to learn is far larger than that represented by our dataset.

\subsection{Expert Human Knowledge Prior}

To capture professional notions of style, we introduce a triplet loss into our architecture. Given a triplet of images $(x_a, x_p, x_n)$ which represent the anchor, positive, and negative images respectively, the objective is to minimize the distance between the anchor and positive instances while maximimizing the distance between the anchor and negative samples in our learned style manifold. In our approach, each image in our portrait dataset was paired with two other randomly chosen portraits from the dataset, and a professional art historian labelled which of the two images was the positive and negative sample. Using these labels as supervision during training, we imbue in the model a professional notion of style, forcing stylistically similar images should be closer together in the learned style subspace, and stylistically dissimilar images to be farther apart.

Typically when employing a triplet loss, triplet mining is done to only extract the hardest of the polynomial number of possible triplets, however, this requires that the label can be automatically determined for any arbitrary triplet. In our approach, triplets require professional labelling. Therefore, we cannot employ traditional triplet mining techniques as
described in \cite{DBLP:journals/corr/HermansBL17,Schroff2015}. Instead we avoid training on triplets that trivially satisfy the triplet constraint by normalizing the loss by only the number of triplets not satisfying the triplet constraints, $N_{+}$. 

We define the triplet loss similarly to that of FaceNet as described in \cite{Schroff2015} in that we also introduce a margin $\alpha$ of separation between the positive and negative samples with respect to the anchor sample. Given the encoded means of the anchor, positive, and negative images respectively as $\mu_a, \mu_p, \mu_n$ we formally define our triplet loss to be:
\begin{align} \label{eq:triplet}
\mathcal{L}_{TRIPLET} &= \frac{\lambda}{\max(N_{+}, 1)}\sum_{i}^N \left[\left\|\mu_a-\mu_p\right\|_{2}^{2}-\left\|\mu_a- \mu_n\right\|_{2}^{2}+\alpha \right]_{+}
\end{align}

such that
\begin{align} \label{eq:N}
N_{+} &= \sum_{i}^N \mathbbm{1} (\left\|\mu_a-\mu_p\right\|_{2}^{2}-\left\|\mu_a- \mu_n\right\|_{2}^{2}+\alpha > 0).
\end{align}

\subsection{Statistical Prior}

We choose a VAE as the base architecture for our model trained with our portrait dataset due to both the statistical prior incorporated in the architecture as well as the interpretable latent space. The statistical prior provides us with a reasonable embedding of style based on expected distributional properties of the space of style representations, particularly since the space we are trying to learn is far larger than that represented by our dataset. Furthermore, the latent style construction of the VAE allows us to generate novel images interpolated between the styles of existing images in our dataset enabling interesting analysis of the learned style space not feasible with most other neural network architectures.

The loss function for a VAE is defined as
\begin{equation}
    \mathcal{L}_{\text{VAE}} = \lambda_1 \mathcal{L}_{KL} + \lambda_2 \mathcal{L}_{RECON} = \lambda_1 D_{KL}\left(p_{\theta}(z | x)\Vert\mathcal{N}(0,1)\right) + \lambda_2 \Vert I-\widehat{I} \Vert^2_2
\end{equation}
where $z$ is the latent space of the network, $x$ is the training data, $\theta$ is the parameters of the network, and $I$ is an input image from the training data with associated reconstruction by the network $\widehat{I}$. In $\mathcal{L}_{\text{VAE}}$, we recall that $\mathcal{L}_{KL}$ imposes a loss on the learned latent space distribution against the assumed prior distribution, the unit standard Gaussian distribution $\mathcal{N}(0, 1)$. This fixed latent space prior is the statistical prior we evaluate for our style representation study.

To improve the reconstruction quality in all of our VAE experiments, we add a super-resolution perceptual loss, first introduced in \cite{Johnson2016Perceptual}. This loss helps circumvent the issues with just using a mean-squared error to enforce high reconstruction quality as discussed in \cite{Mathieu2016}.

\begin{figure*}[t]
    \centering
    \includegraphics[scale=0.13]{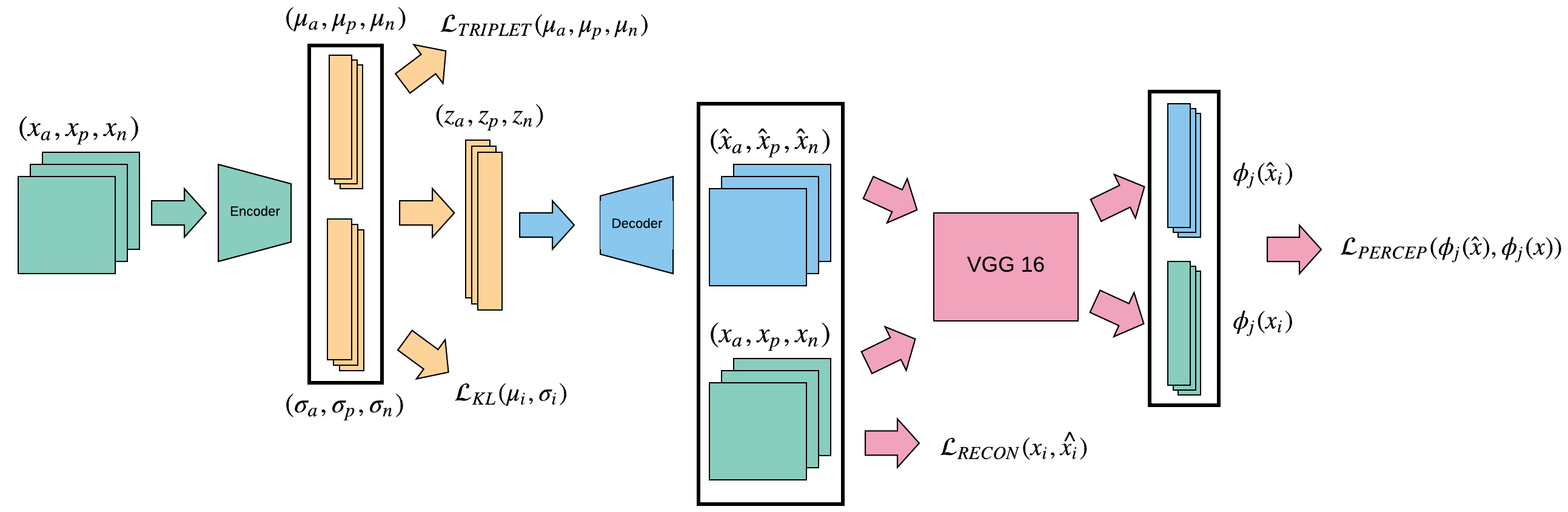}
    \caption{Model architecture for our VAE including the triplet loss. All the triplets $(x_a, x_p, x_n)$ are passed to the network, and the triplet loss is applied in the latent space of the VAE. The VAE reconstructions are also passed into a VGG-16 network pretrained on ImageNet so that the VGG-16 activations can be used to compute a perceptual loss to enhance the reconstruction quality of the network.}
    \label{fig:vae}
    \vspace{-1em}
\end{figure*}

To incorporate the expert human knowledge prior with our VAE architecture, we simply add the triplet loss specified in equations \ref{eq:triplet} and \ref{eq:N} to the total VAE loss optimized jointly. Figure \ref{fig:vae} visualizes this full model architecture.

\subsection{Photo Realism Prior}

Neural networks designed for analysis of photo collections have been shown to be able to learn representations of low level image features such as color, edges, and texture which are useful outside of the datasets used in training \cite{Johnson2016Perceptual}. These features when derived from digital paintings are components of artistic style, though they do not explicitly seem to encode higher level style characteristics. We study the ability of networks given prior information regarding these low level image characteristics via pretraining on ImageNet, a dataset of photographs, to extract meaningful representations of artistic style from our portrait dataset. We note that by leveraging datasets of photographs as a meaningful prior for artistic style analysis, we can partially address the challenge of limited digital datasets of artwork available for computational analysis.

We produce the style embeddings for our analysis from a VGG-16 model \cite{Simonyan2015} pretrained on ImageNet. We remove the final fully connected classification and sigmoid layers to yield a $4096$ length feature embedding. We expect that these feature embeddings will be strongly correlated with lower frequency characteristics such as the content of the image.

To incorporate the expert human knowledge prior with this photo realism prior, we remove the final classification and sigmoid layers again from the pretrained VGG-16 model to yield a perceptual $4096$ length feature embedding. We then add two additional linear layers to generate the $1024$ length style embedding. Thus, the feature activations from our digital portrait data passed through the frozen, pretrained VGG-16 model are used as input to the described style encoder. We train this style encoder on our portrait dataset with the triplet loss function, $\mathcal{L}_{TRIPLET}$, calculated with the generated style embeddings for each triplet image in place of the means in equations \ref{eq:triplet} and \ref{eq:N}.

\section{Representation Analysis}

%I think one of the differences that makes this paper a unique contribution is also its approach to metric learning--an attempt to adopt some of the descriptive terms of art history itself--to think about form, content, color, etc. as part of style. 

We now present the results of our study on style representations of our portrait dataset. Implementation details and source code to reproduce our work is publicly available on our project GitHub\footnote{https://github.com/learning-portrait-style/learning-style}. In our experiments, we use the following four models presented in Section \ref{sec:priors} which use combinations of each of the three information priors: expert human knowledge, statistical, and photo realism.

\begin{enumerate}
    \item VAE (Statistical Prior)
    \item VAE + Triplet Loss (Statistical + Expert Human Knowledge Priors)
    \item VGG-16 (Photo Realism Prior)
    \item VGG-16 + Triplet Loss (Photo Realism + Expert Human Knowledge Priors)
\end{enumerate}

For all of our models, we perform dimensionality reduction on the learned style representations to analyze their placement in a projected 2D feature space relative to each other with clusters colored by artist label. We observe the VGG-16 + Triplet Loss model provides priors which yield the most sensical artistic style representations. We then visualize style-based activation maps for the VGG-16 models to observe the impact of the triplet loss on the emphasized positive and negative examples of relative style between images within a triplet from our test data. For the VAE models, we demonstrate the impact of the triplet loss on providing stylistic meaning to the latent space of the model by interpolating between latent image reconstructions. We observe novel images generated in intermediate styles throughout this interpolation only when the triplet loss is included in training the network. Finally, we use the learned style features from each of the four models to perform zero-shot classification of the source artists. We find that the features learned by the VGG-16 + Triplet Loss model have the most predictive power.

\subsection{Style Embeddings Clustered by Artists} \label{sec:tsne}

\begin{figure}[h!]
\centering
\begin{subfigure}{.9\linewidth}
    \centering
    \includegraphics[scale=0.24]{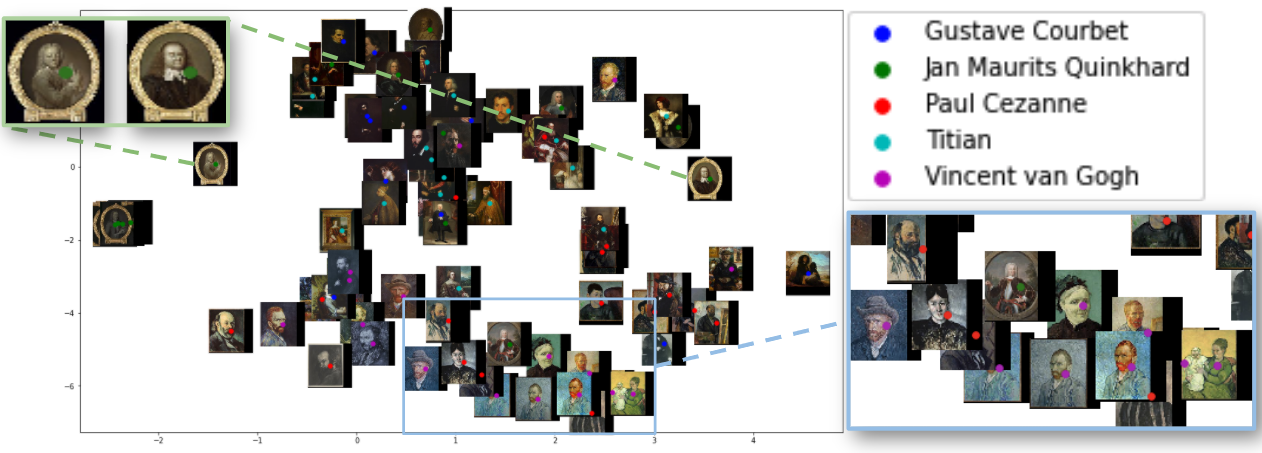}
    \caption{VAE}
    \label{fig:notriplet}
\end{subfigure}
\begin{subfigure}{.9\linewidth}
    \centering
    \includegraphics[scale=0.27]{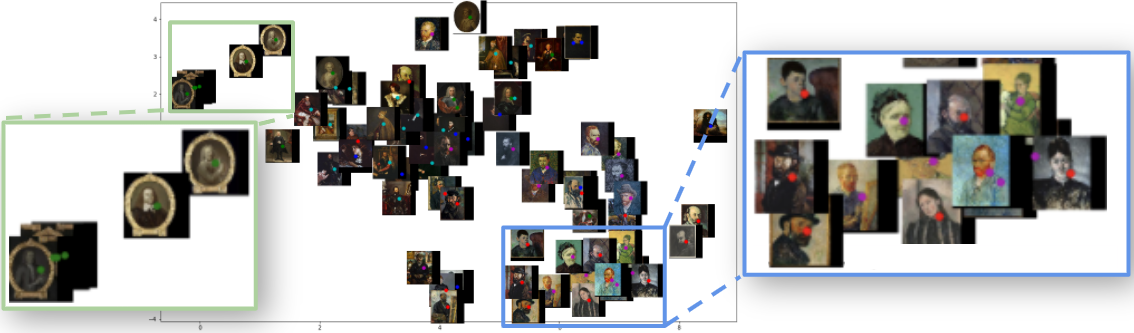}
    \caption{VAE + Triplet Loss}
    \label{fig:vae_triplet}
\end{subfigure}
\begin{subfigure}{.9\linewidth}
    \centering
    \includegraphics[scale=0.27]{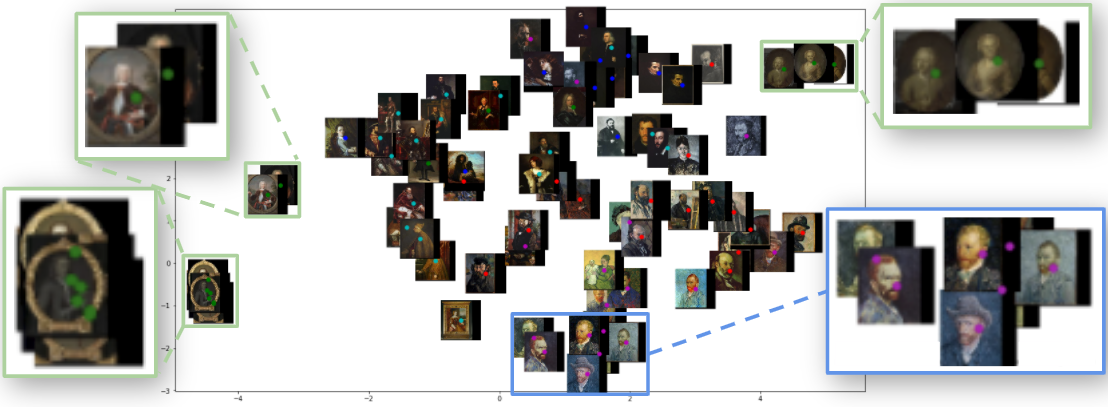}
    \caption{VGG-16}
    \label{fig:vgg}
\end{subfigure}
\begin{subfigure}{.9\linewidth}
    \centering
    \includegraphics[scale=0.27]{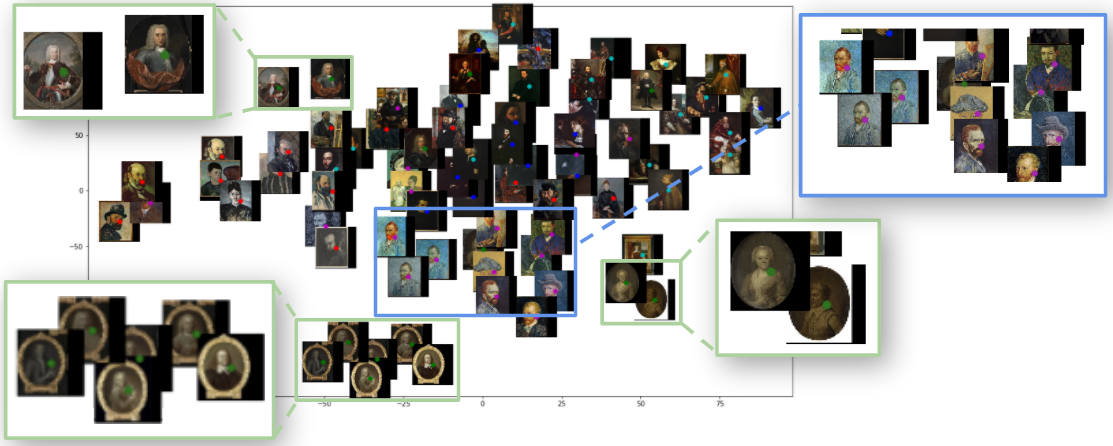}
    \caption{VGG-16 + Triplet Loss}
    \label{fig:vgg_vae}
\end{subfigure}
\caption{2D t-SNE projection of test set embeddings from each model.}
\label{fig:tsnes}
\end{figure}

To analyze the style embeddings produced for each model, we use PCA with t-SNE \cite{maaten2008visualizing} to project our representations from each model into a two dimensional space. We note that we use these dimensionality reduction methods together to suppress noise since the number of feature dimensions is very high as proposed by \cite{van2014accelerating}. We then visualize the points from the five most represented artists in the portrait test data in Figure \ref{fig:tsnes} and interpret the latent clusters in the context of professional analysis of the artistic style in each portrait. In each t-SNE plot in Figure \ref{fig:tsnes}, an expanded view is provided of the image clusters on which our discussion focuses in the text.

\textbf{Statistical Prior (VAE, Fig.~\ref{fig:notriplet}) }
%\textbf{VAE} \textit{Figure \ref{fig:notriplet} : Statistical Prior}\\
With a statistical prior alone the model exhibits non-trivial clustering within the latent space. On the left border of the plot, many of the Jan Maurits Quinkhard paintings are clustered together tightly, indicating strong stylistic similarity between the framed portraits. Along the vertical t-SNE direction, there exists a transition from the austere portraiture of Corbet at the top of the plot to the colorful portraits like those of Van Gogh. 

However, along the horizontal t-SNE direction, there is no apparent cohesive stylistic interpretation as there was for the vertical t-SNE direction. Additionally, some portraits that are clearly stylistically related, such as from Jan Maurits Quinkhard's miniatures appear on opposing sides of the plot. These significant gaps between the learned style representation and well-accepted human style interpretations lead us to doubt the useful of a statistical prior alone in enforcing meaningful embeddings of artistic style.

%%% ORIGINAL WILL TEXT START %%%%%
%NT: The organization of the Y-axis makes sense (with an outlier, maybe two), but the breadth of the X-axis is confusing, mostly because similar paintings end up on either extreme.
%%% ORIGINAL WILL TEXT END %%%%%

\textbf{Statistical and Expert Human Knowledge Priors (VAE + Triplet Loss, Fig.~\ref{fig:vae_triplet}) }
The addition of model supervision via our professionally labelled triplets to the latent space optimization of our VAE translates into meaningful semantic differences for the style representations visualized in our t-SNE plots. For instance, whereas the model trained with just the statistical prior had difficulty clustering the Jan Maurits Quinkhard miniatures, this model was able to successfully group the miniatures together in addition to distinguishing between different miniature frame textures. 

Similarly to VAE model without the triplet loss, the Van Gogh paintings were grouped together, along with some of Paul Cezanne's works. This grouping marks a distinction between the the works of Van Gogh and Cezanne as opposed to the more austere works of Courbet. This plot also conveys more apparently the bizarre nature of Courbet's painting near the right border of the plot, which is in stark contrast to the rest of the present works in both content and texture.

%%% ORIGINAL WILL TEXT START %%%%%
% NOTE FROM SADAT: : This first paragraph was from the second plot generated -- I'm not sure how much of it we can still use.
% B I like less. At first blush, it works together as an art piece. Each painting seems similar to the one next to it, but as a comment on art history, it is too jumbled and disperse to make sense. 
% Courbet and Cezanne are much too spread out, and where they show up seems to be determined more by arbitrary aspects of their composition or low-level features--like the pose or size of the sitter, or the choice of color palette--than an actual insight into style. That said, it isn't horrible, just messy.

% R and T: Are both more interesting to me. They each get a couple things, maybe three, very wrong, in my opinion, but are richer and more interesting, precise, and nuanced. I was initially really drawn to T because of the way it sorted the Quinkhard-- it separated the miniatures well, but it also put his more naturalistic paintings nearer the Y-axis--how it place Courbet's bizarre portraits far to the right--and the way it grouped the Post-Impressionists together without being too caught up with the artist themselves.

\textbf{Photo Realism Prior (VGG-16, Fig.~\ref{fig:vgg}) }

With the strong photo realism prior in our VGG-16 model, we see a more systematic grouping of the artwork, primarily driven by lower frequency, perceptual similarities between images. For instance, more groups emerge among the Jan Maurits Quinkhard paintings. As before, the miniatures are grouped tightly together, with less of a distinction between the frame textures as was present in the previous model where supervision from our triplet labels was introduced in training. However, unlike before other Quinkhard groupings emerged such as the group of portraits in the top right and middle left of the image which bear strong content similarities within their respective group but differ greatly from the rest of Quinkhard's works. 

Additionally, whereas the VAE models grouped the Van Gogh paintings among Cezanne's paintings due to their textural similarities, they are now separated into their own distinct clusters, implying a more nuanced view of each artist's style. Cezanne's portraits are also spread out greatly among the horizontal t-SNE axis, a defensible position due to his style developing over a much longer period, often changing significantly during this time.

\textbf{Photo Realism and Expert Human Knowledge Priors (VGG-16 + Triplet Loss, Fig.~\ref{fig:vgg_vae}) }
The combination of high and low level feature sorting in the VGG-16 model including the triplet loss resulted in a very visually orderly embedding with intuitive relative spacing across the style space. The clusters of Van Gogh and Cezanne art are close to each other in many portions of the plot, but with some clear placement distinctions corresponding to the connection between their respective artistic styles. Van Gogh and Cezanne's brush strokes are very different even if they both file under Post-Impressionism. 
\begin{wrapfigure}{l}{0.6\textwidth}
    \vspace{-2em}
    \centering
    \includegraphics[scale=0.25]{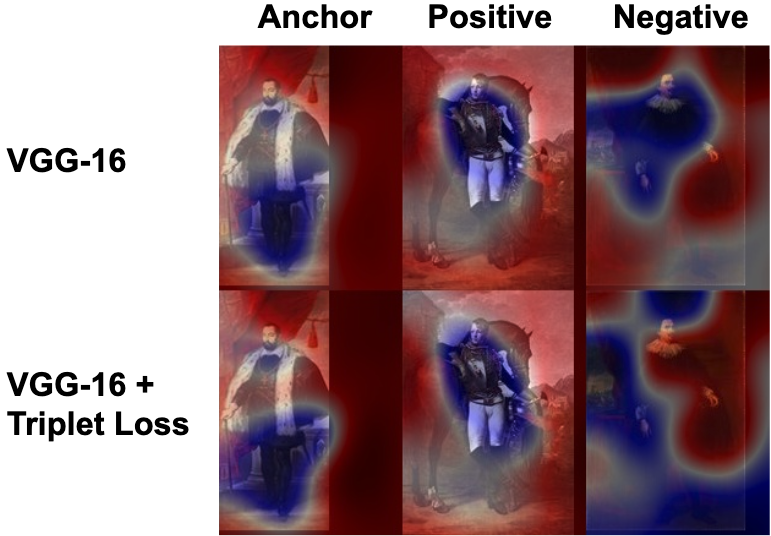}
    \caption{Activation maps which highlight regions the VGG-16 and VGG-16 + Triplet Loss models each learned to be of stylistic importance when computing a triplet loss on an image from our portrait test set.}
    \label{fig:guided_backprop}
    \vspace{-2em}
\end{wrapfigure}
Cezanne has a geometric brush stroke that is very contained and precise, Van Gogh has a dynamic brush stroke that works very well with the whole. As you step closer to a Van Gogh, you become entranced by its texture. As you step close to a Cezanne, the painting seems to fall apart, it's illusion of depth and form crumbles to reveal, not thick, impassioned brush strokes like in Van Gogh's work, but flat patches of color. Cezanne's portraits are again more spread out, reflecting his changes in style over time.

Furthermore, Titan paintings are less dispersed throughout the plot than Van Gogh or Cezanne, appropriate reflecting his more constrained style choices. We also see the desirable attributes of the Quinkhard clusters observed with the previous models in this plot as well. Overall, this model aligns most closely with what we understand to be a meaningful style interpretation of these paintings and the work of the associated artists from well-accepted art historical analysis.

\subsection{Visualizing Style-Based Activation Maps} \label{sec:act_maps}

Producing style embeddings through our models is useful to art historians, however, generating style embeddings alone does not answer the essential questions of ``how" and ``why" these style embeddings were produced. Answering these questions is crucial as it allows art historians to not only spot flaws in the models' understanding of style, but also potentially gain insights into relationships of style between images that may not be traditionally observed otherwise.

To answer these questions, we seek to generate activation maps which highlight regions of stylistic importance within an image, and compare these maps among the triplets to understand relationships of style. To generate these activation maps we utilize a similar approach to Grad-CAM, as described in \cite{Selvaraju2017} which uses the gradients produced from the network loss function to produce a coarse activation map denoting regions of importance in the image driving network weight optimization. We compute activation maps computed from a triplet loss calculation on test data for both VGG-16 models. These maps are displayed in Figure \ref{fig:guided_backprop} for a triplet from our portrait test data. 

\begin{wrapfigure}{r}{0.6\textwidth}
    \vspace{-2em}
    \centering
    \includegraphics[scale=0.27]{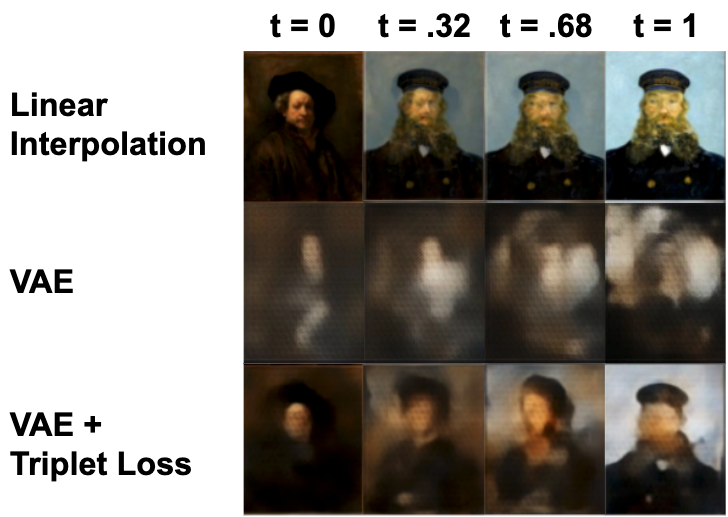}
    \caption{Decoded interpolated style embeddings with interpolation factor $t$ of style embeddings $z_{src}$ ($t=0$) and $z_{tgt}$ ($t=1$)}.
    % \caption{Reconstructions from our VAE of portraits in the test set. The first row is the linear interpolation between the source and target test set images with factor $t$. The second row is the style conditioned reconstruction from the VAE trained \textbf{without} the triplet loss. The third row is the style conditioned reconstruction from the VAE trained \textbf{with} the triplet loss.}
    \label{fig:reconstructions}
    \vspace{-2em}
\end{wrapfigure}

%\subsubsection{Analyzing expert human knowledge impact on activation maps:} TODO WILL: STYLE ANALYSIS
We note a significant similarity in how the activation maps appear for the models trained with and without the triplet loss. This similarity is derived from the photo realism prior induced by pretraining each VGG-16 model on the ImageNet dataset. This pretraining biases both models towards interpreting regions of the images which appear frequently in ImageNet, such as the background trees and greenery. Those regions are highly activated in both the anchor and positive images in Figure \ref{fig:reconstructions}. However, there is a stark difference between the two models' activation maps for the negative triplet image. The triplet loss emphasizes the strong stylistic ramifications the subject's pose and clothing rather than focusing on the background elements more common in our training datasets. This observation lends evidence that the model understands and can prioritize higher frequency characteristics such as garment texture in addition to lower frequency characteristics such as natural background elements.

\subsection{A Continuous Space of Style}

One of the benefits of utilizing a variational approach to learning a style manifold is that the model stochasticity improves the smoothness of the underlying learned style manifold in contrast to purely deterministic approaches which can be prone to over-fitting. By encoding a smooth style space, as well as learning a style conditioned generator, we gain the ability to visualize sampled points along the manifold, allowing for style-conditioned novel image generation. 

In Figure \ref{fig:reconstructions} we consider the source portrait of ``Self-Portrait" by Rembrant van Rijn and the target portrait of ``Portrait of Postman Roulin", painted by Vincent Van Gogh, two portraits that are stylistically quite dissimilar. In the VAE model without use of our triplet loss, the interpolated style embeddings do not contain meaningful information. In our VAE model including the triplet loss, we observe two novel portraits that represent neither the source nor the target images, implying the latent space of this model contains an understanding of style beyond that needed for reconstruction tasks.

%TODO WILL: STYLE ANALYSIS

% where by sampling interpolating points between a source and target style embedding we produce 

% Explain latent interpolation.

% 3 rows of image reconstructions with interpolation between images with commentary (find will's notes on this): 1) image averaging, 2) vae latent interp, 3) vae + triplet latent interp

% We show through this approach the ability of the model to not only reconstruct the input data but also be able to generate novel artwork, as demonstrated in Figure \ref{fig:reconstructions}. Though novel style-conditioned portrait synthesis confirms some of the intended behavior of our trained model, our primary purpose of this work is to learn a meaningful style distribution of art portraiture. Thus, our primary experiments focus on the analysis of the clustering in this learned style space as well as analyzing the difference in feature activations from the learned models.

\subsection{Zero-Shot Classification from Style Embeddings}

%Style attributes for portraits can range from low frequency characteristics such as the deliberate choice of portrait content to higher frequency textural characteristics such as brush stroke and technique. In this work we hypothesize that by the model learning how to represent style, the model also implicitly learns how to identify style attributes unique to a particular artist.
\begin{wrapfigure}{r}{0.5\textwidth}
    \vspace{-2em}
    \centering
    \includegraphics[width=0.48\textwidth]{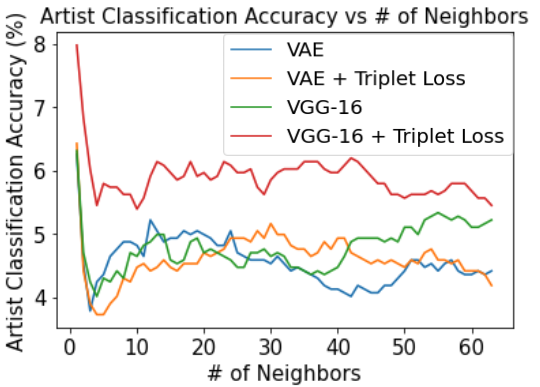}
    \caption{Artist classification accuracy for different learned style representations.}
    \label{fig:artist_class}
\vspace{-2em}
\end{wrapfigure}

We claim in our work that good style representations should be identifying of artistic metadata we know to separate closely along the lines of clear style distinctions. To demonstrate this association, we use the style features learned by each of our four models to classify artists using k-nearest neighbors classification based on Euclidean distances between style embeddings. The classification accuracies are plotted in Figure \ref{fig:artist_class}, against the number of neighbors used when fitting the classifier. We observe that the features learned by the VGG-16 model with the triplet loss has the highest classification accuracy. This model also had the most stylistically intuitive features in our analyses in Section \ref{sec:act_maps} and \ref{sec:tsne}, supporting our claim.
%To test this hypothesis, we generate style embeddings on the entire training set and fit a K-Nearest Neighbor classifier using the euclidean distance between style embeddings. The classification accuracies are plotted in figure \ref{fig:artist_class}, against the number of neighbors used when fitting the classifier.
Due to the low number of samples per artist class, it is not surprising the optimal number of neighbors for all models was one. In all models, we note a non-trivial classification accuracy. Our portrait data test set contains 943 classes, some of which not in the training set, representing a baseline trivial accuracy of just .106\%.

\section{Conclusion}

In this work, we presented a study of style representations learned by neural networks incorporating different priors. In particular, we analyzed the impact of priors from distributional assumptions, expert human knowledge, and photo realism on the representations of style features learned by neural networks. To facilitate this research, we also presented the first large-scale dataset of portraits prepared for computational analysis. In this dataset, we introduced triplet labels for style similarity annotated by a professional art historian. We observed that style representations aligning most closely to art historical analysis with the most predictive power for the source artists came from using priors for photo realism with expert human knowledge given by supervision from our triplet labels. In future work, we intend to continue the direction of this study by expanding our dataset to include additional metadata fields and make new dataset with content focuses beyond portraiture. We will then be able utilize this data to extend our investigation of style representations. We plan to study the relation of our continuous style representation to discrete labels like Wölfflin concepts. Additionally we intend to explore other other generative models that learn rich underlying latent spaces such as the generative flow model presented in \cite{Kingma2018}.

\clearpage
\bibliographystyle{splncs04}
\bibliography{egbib}
\end{document}